\documentclass[11pt]{article}

\usepackage[]{acl}

\usepackage{times}
\usepackage{latexsym}
\usepackage{amsmath} 
\usepackage{amssymb}
\usepackage{booktabs} 
\usepackage{multirow}
\usepackage{makecell}
\usepackage{cuted}
\usepackage{caption}
\usepackage{tabularx}

\usepackage{graphicx}
\usepackage{subcaption}
\usepackage{caption}
\usepackage{comment}

\usepackage{graphicx}
\usepackage{subcaption}  
\usepackage[font=small, labelfont=bf]{caption} 

\usepackage[T1]{fontenc}
\usepackage{enumitem}

\usepackage[utf8]{inputenc}

\usepackage{microtype}

\usepackage{inconsolata}

\usepackage{graphicx}

\usepackage{cuted}
\usepackage{capt-of}

\usepackage[table]{xcolor}

\usepackage{CJKutf8}
\newcommand{\pchn}[1]{\protect\begin{CJK*}{UTF8}{gbsn}#1\protect\end{CJK*}}

\usepackage[T1,T5]{fontenc}
\usepackage[vietnamese,turkish,english,shorthands=:!]{babel}

\definecolor{C1}{RGB}{104,132,147}
\definecolor{C2}{RGB}{172,194,226}
\definecolor{C3}{RGB}{91,98,101}
\definecolor{C4}{RGB}{131,157,160}
\definecolor{C5}{RGB}{235,238,240}

\title{When Languages Disagree: Self-Evolving Multilingual LLM Judges}

\author{Xiyan Fu and Wei Lu \\
  Nanyang Technological University\\
  \texttt{\{xiyan.fu, wei.lu\}@ntu.edu.sg} \\}

\begin{document}
\maketitle
\begin{abstract}
Multilingual LLM-as-a-judge is widely used to evaluate model outputs across languages, but suffers from cross-lingual inconsistency \citep{fu-liu-2025-reliable}. Existing methods typically treat this inconsistency as noise and mitigate it through voting or aggregation. In this work, we instead show that multilingual inconsistency can provide complementary evaluation signals. 
Our oracle analysis finds that sampling judgments across languages yields a higher performance upper bound than single-language judging, indicating that different languages potentially include complementary judgments. Motivated by this finding, we propose SEMJ, a self-evolving multilingual judge that leverages cross-lingual inconsistency for iterative refinement. SEMJ constructs multilingual variants of each input, collects independent judgments and rationales, and feeds inconsistent outputs back for self-reflection and re-evaluation. Experiments on multiple benchmarks show that SEMJ consistently outperforms voting and reflection baselines in both accuracy and cross-lingual consistency. Further analysis shows that inconsistency triggers useful re-evaluation, which improves judgment quality.
\end{abstract}

\section{Introduction}

\begin{figure}
    \centering
    \includegraphics[width=1\linewidth]{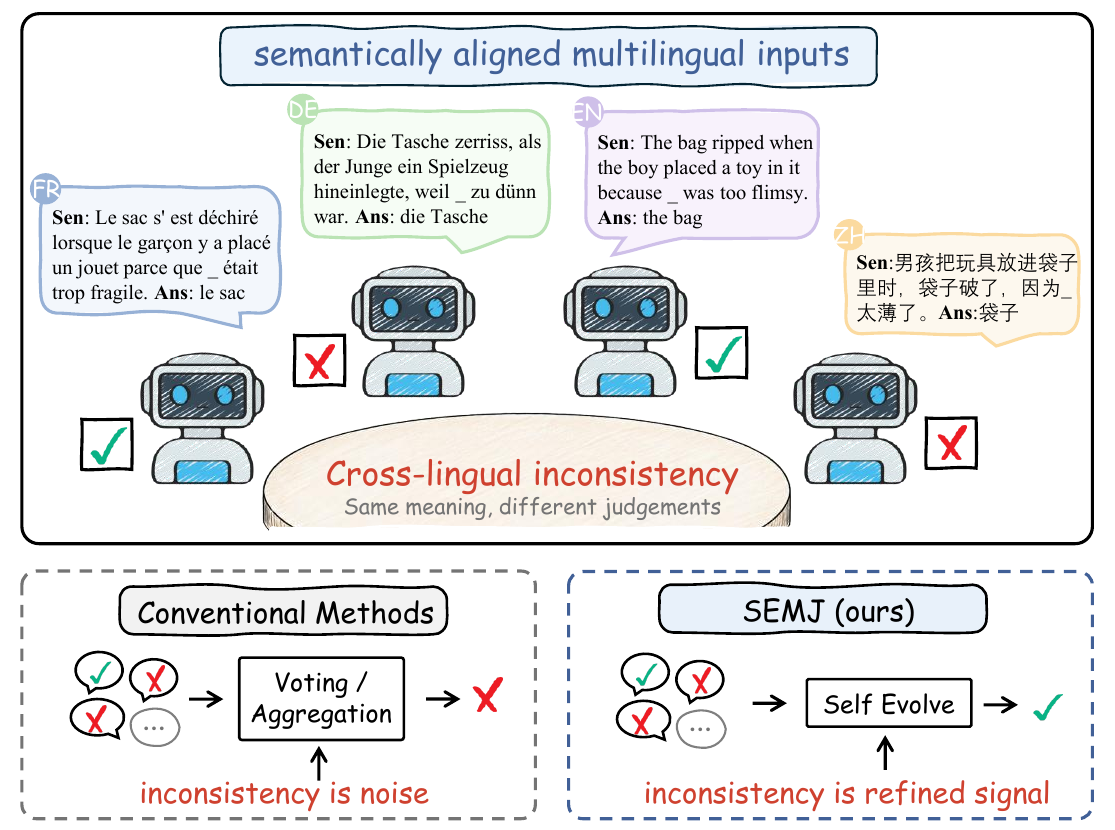}
    \caption{\textit{top}: multilingual judge inconsistency, where semantically aligned multilingual inputs (\underline{Sen}tence + \underline{Ans}wer) elicit different judgments from the same judge across languages. \textit{bottom}: i) conventional aggregation methods, which treat inconsistency as noise and reduce it to a single vote; ii) our approach SEMJ, which treats inconsistency as a refinement signal for self-evolving judgment refinement.}
    \label{fig:intro}
\end{figure}

Multilingual LLM-as-a-judge has become a widely adopted paradigm for evaluating model predictions across languages due to its strong multilingual semantic understanding and flexible evaluation capability \citep{zheng2023,GU2026101253}. Despite its success across diverse tasks, recent studies have revealed cross-lingual inconsistency, where semantically equivalent inputs in different languages can lead to different evaluation outcomes \citep{qi-etal-2023-cross, fu-liu-2025-reliable, wang-etal-2025-lost-multilinguality}. Figure~\ref{fig:intro} illustrates an example. This inconsistency raises serious concerns about the reliability of multilingual judge, as evaluation results may depend on language choice rather than model predictions.

To mitigate this issue, existing approaches often generate multiple candidate judgments through repeated sampling and aggregate them via majority voting \citep{haldar-hockenmaier-2025-rating} or likelihood aware aggregation \citep{wang-etal-2025-improving-llm-judge, wang2025trustjudge}. These methods implicitly treat inconsistent judgments as undesirable noise that should be relieved. Rather than viewing such cross-lingual inconsistency as purely harmful, we ask whether it can provide complementary evaluation signals elicited by different languages. To answer this question, we conduct a preliminary oracle analysis (Section \ref{sec:complementary}), and find that multilingual sampling consistently achieves higher oracle performance than repeated decoding within a single language. This finding suggests that multilingual variants expose more potentially correct judgments.

Motivated by this observation, we propose \textbf{SEMJ}, a \textbf{S}elf-\textbf{E}volving \textbf{M}ultilingual \textbf{J}udge framework that leverages cross-lingual inconsistency as a refinement signal rather than relieving it through voting. For each evaluation instance, SEMJ first constructs multiple semantically equivalent variants in multilple languages. These variants are then independently evaluated by the same multilingual judge, producing corresponding judgments and rationales across languages. The resulting cross-lingual inconsistencies, together with their associated rationales, are subsequently fed back to the judge model to trigger iterative self-reflection and re-evaluation. Through multiple rounds of refinement, SEMJ is expected to progressively improve both evaluation accuracy and cross-lingual consistency without requiring additional supervision.

Experiments across multiple benchmarks and model backbones show that SEMJ consistently outperforms existing voting and reflection based baselines in both evaluation accuracy and cross-lingual consistency. Notably, the improvements are more pronounced in low resource languages, while remaining gains are observed in medium and high resource settings, suggesting that cross-lingual inconsistency is not solely driven by high-to-low resource transfer. Beyond empirical improvements, we further investigate how multilingual inconsistency contributes to judge refinement. We conduct sparse auto-encoder analysis, where we find that inconsistency often triggers an internal re-evaluation process, encouraging the judge to reconsider and revise its initial judgments. 
More importantly, we find that the benefit of this process mainly comes from corrective rationales associated with correct predictions, which are more specific and better aligned with the evaluation rubric.

Our main contributions are as following\footnote{Our code is available at \url{https://github.com/xiyan524/semj-}}:
\begin{itemize}[noitemsep, topsep=5pt]
    \item We identify cross-lingual inconsistency as a useful signal in multilingual LLM-as-a-judge, and show via oracle analysis that multilingual settings expose more potentially correct judgments than single-language sampling.
    \item We propose a Self-Evolving Multilingual Judge framework that leverages cross-lingual inconsistency for iterative refinement. Experiments demonstrate its effectiveness across multiple benchmarks and model backbones.
    \item We analyze the working mechanism of SEMJ and find that cross-lingual inconsistency acts as a trigger for internal re-evaluation, where more detailed and correct rationales contribute to improved judgment quality.
\end{itemize}

\section{Related Work}
\paragraph{LLM-as-a-Judge}
Early work primarily focuses on prompt design to elicit evaluation reasoning from pretrained LLMs, such as incorporating carefully crafted examples into prompts \citep{lin-chen-2023-llm, fu-etal-2024-gptscore}. 
Beyond prompt design, another line of work improves LLM evaluators via task-specific tuning. Prior work constructs synthetic datasets for fine-tuning \citep{xie2025sorrybench}, or relies on human-annotated data \citep{kim2024prometheus, zhu2025judgelm}, while others train evaluators using preference signals \citep{wu2024meta, wang2024pandalm}. To avoid human-crafted signals, recent works explore self-evolving LLM judges by leveraging model-generated signals. Existing approaches either use self-generated references to enhance evaluation at inference time \citep{lin-etal-2025-judge}, or construct training supervision from the model’s own judgments and rationales \citep{wang2024self, trivedi2024self}, with some extensions introducing meta-evaluation signals to further refine judgment quality \citep{wu-etal-2025-meta}. These approaches rely on a single model to perform judgment. More recently, LLM judges have been extended to multi-judge settings that aggregate signals from multiple models, through cooperation \citep{liang-etal-2024-abseval}, aggregation \citep{verga2024replacing}, or debate \citep{zhao-etal-2025-auto}.

These approaches primarily rely on internal signals derived from a single language or model behavior. In contrast, we exploit cross-lingual inconsistency as a self-evolving signal, resulting in more robust and reliable multilingual LLM judges.

%

%




\paragraph{Self-Evolution in LLMs} 
Recent work proposes to treat LLMs as adaptive systems which can be iteratively improved from their own feedback \citep{fang2025comprehensive,xiang2026systematic,gao2026a}. One well-known category is \textit{inference based evolution} that uses additional computational resources within the inference process to enhance reasoning performance \citep{dong2024survey}. Most works use parallel computing to provide more broad solution coverage to avoid suboptimal inference, such as collecting a diverse set of results from sampling \citep{wang2023selfconsistency} or various LLMs \citep{jiang-etal-2023-llm}. Others leverage iteratively alternate between generation and revision, such as using memory of failures \citep{shinn2023reflexion} or incorporating external tools for feedbacks \citep{gou2024critic}. To avoid large inference cost, the other category \textit{training-based evolution} proposes to achieve permanent capability internalization by parameter updates. They generate multiple candidates and derive preference signals via model-based evaluation, either through self-scoring \citep{yuan2024selfrewarding} or debate-based comparative judgment \citep{srivastava-etal-2025-debate}, followed by preference optimization \citep{rafailov2023direct}.

Our work belongs to \textit{inference based evolution}. In contrast, we use \emph{cross-lingual inconsistency} as the  evolution signal. This design yields a more robust multilingual judge.

\begin{figure}
    \centering
    \includegraphics[width=0.8\linewidth]{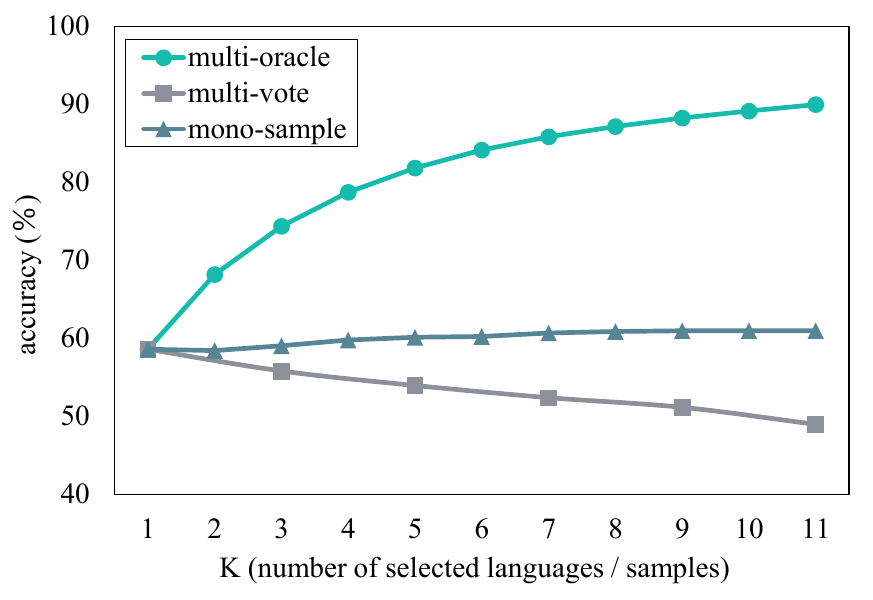}
    \caption{Performance of complementarity in multilingual sampling on XCOPA. We compare multilingual oracle (multi-oracle), multilingual majority voting (multi-vote), and monolingual $K$-sampling (mono-sample) under different values of $K$, i.e, the number of selected languages or samples. }
    \label{fig:compli}
\end{figure}



\begin{figure*}
    \centering
    \includegraphics[width=1\linewidth]{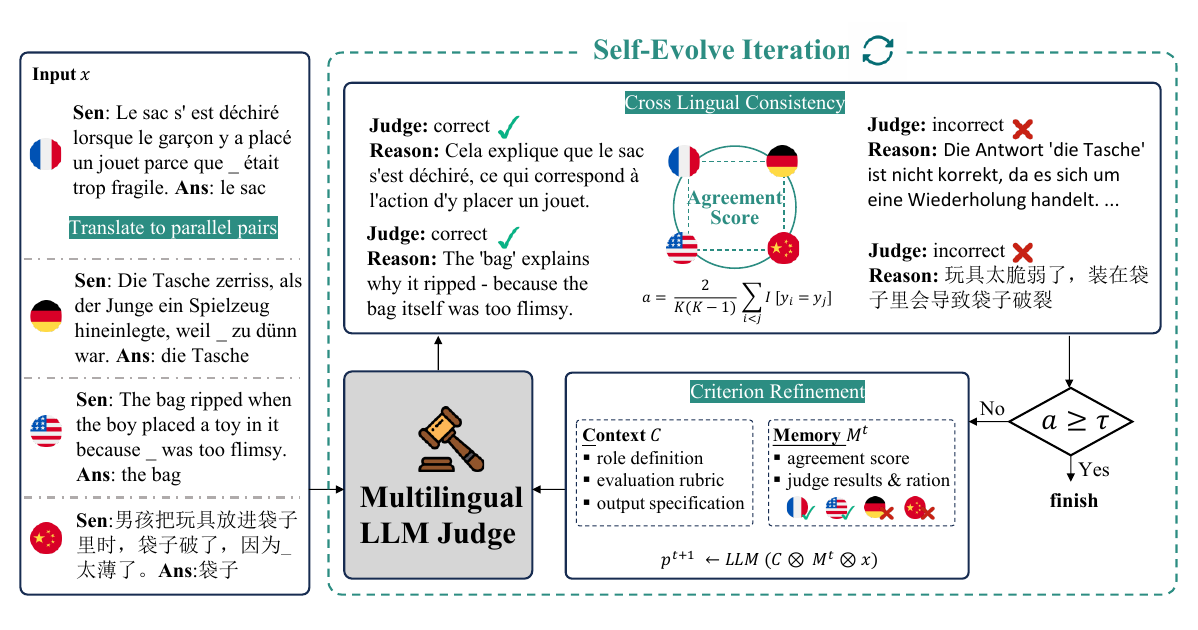}
    \caption{Overview of the inconcsistency-driven self evolution multilingual LLM judge. Given an input sample $x$, we first construct semantically equivalent multilingual variants. They are sent to LLM judge to independently produce judgments and rationales. The cross-lingual agreement among these judgments is then used as a consistency signal to guide iterative criterion refinement. If the agreement score does not meet the threshold, the accumulated feedback history is used to update the judging criterion for the next iteration, otherwise, the iteration is terminated.}
    \label{fig:semj}
\end{figure*}

\section{Beyond Noise: Complementarity in Multilingual Sampling}
\label{sec:complementary}
\citet{fu-liu-2025-reliable} show that multilingual LLM judges often exhibit notable inconsistencies across languages. 
While typically viewed as reliability failures, these inconsistencies suggest that semantically equivalent inputs can elicit different judgments. This raises a diagnostic question: whether such inconsistencies are merely noise or instead contain complementary correctness signals?


To answer this question, we conduct a diagnostic oracle analysis to estimate whether cross-lingual differences contain exploitable correctness complementarity. Specifically, we design a multilingual oracle (multi-oracle) setting to measure the upper-bound accuracy achievable by combining judgments from different languages. For each example, we obtain their $K$ language judgement independently. This example is considered correctly solved if any of the $K$ language specific judgments is correct. 
Figure \ref{fig:compli} illustrates results on the XCOPA dataset.\footnote{Additional results on other benchmarks are provided in Appendix \ref{app:complimentary}.} The green curve shows the oracle accuracy consistently increases as the number of languages $K$ grows, indicating a strong potential benefit from multilingual aggregation. 
To further understand where this gain comes from, we compare against two baselines: i) multilingual majority voting (multi-vote), which aggregates one judgment from $K$ languages; and ii) mono-lingual $K$-sample (mono-sample), which draws $K$ independent samples from a single language and applies the same oracle-style success criterion to control for sampling effects.

Figure \ref{fig:compli} shows the multi-oracle consistently outperforms baselines, with the gap widening as $K$ increases. 
The early saturation of mono-sample suggests that repeatedly sampling within a single language provides limited additional correctness information, whereas sampling across languages introduces more complementary signals.
Meanwhile, the inferior performance of multi-vote indicates that these signals are not fully captured by simple majority aggregation.
\textit{Overall, the results suggest that multilingual inconsistency may encode complementary correctness signals, and effectively leveraging them requires more carefully designed mechanisms.}

\section{Self-Evolved Multilingual Judge}

\subsection{Preliminary}
LLM-as-a-Judge \citep{zheng2023,GU2026101253,li-etal-2025-generation} uses powerful LLMs to evaluate model predictions via natural language instructions. Formally, the judging process can be written as $p \leftarrow {\rm{\tt LLM}}(C \otimes x)$, where $x$ is the prediction to be evaluated, $C$ is the evaluation context, and $p$ is the final judgment. The context $C$ typically includes three components: (i) a \textit{role definition}, which specifies the role of the judge model; (ii) an \textit{evaluation rubric}, which defines the assessment criteria; and (iii) an \textit{output specification}, which constrains the format of the returned judgment. In this work, we adopt a pointwise setting, where the judge evaluates each prediction individually. The judge is required to output a binary judgment, along with a corresponding rationale. Table \ref{app:prompt} in Appendix \ref{exp:prompts} provides a detailed example.

\subsection{Judgment Principle Elicitation}
\label{judging principle}

Our method improves judgment quality through iterative self-evolution, illustrated in Figure \ref{fig:semj}. We are motivated by leveraging inconsistencies across semantically aligned samples as feedback signals. 
In the following, we show how the feedback signal is constructed in a single iteration, and then introduce how the criterion is refined across iterations.

\paragraph{Single Iteration}
In each iteration, for each sample $x$, we construct $K$ semantically equivalent variants in different languages, denoted as $\{x_1, x_2, \dots, x_K\}$. The variants are generated through translation by the judge model itself.
We then ask the multilingual LLM judge to evaluate each language variant independently under the current judging criterion, yielding a set of binary judgments $\mathcal{Y}^{(t)}=\{y_1^{(t)}, \dots, y_K^{(t)}\}$ at iteration $t$. Since these variants express the same underlying semantic content, a robust multilingual judge is expected to produce consistent judgments across languages. To quantify this consistency, we measure the agreement among the predicted labels for all language variants of the same sample. Specifically, we define the agreement score at iteration $t$ as:
\begin{equation}
a^{(t)} = \frac{2}{K(K-1)} \sum_{1 \le i < j \le K} \mathbb{I}\left[y_i^{(t)} = y_j^{(t)}\right],
\end{equation}
where $\mathbb{I}[\cdot]$ is the indicator function. This score computes the proportion of language pairs that receive identical judgments, regardless of whether the prediction label is positive or negative. A higher agreement score indicates that the current judge makes more consistent decisions across semantically aligned multilingual variants, while a lower score suggests stronger cross-lingual inconsistency.

Beyond the binary labels, we also retain the corresponding rationales $\{r_k^{(t)}\}_{k=1}^{K}$ as auxiliary evidence. While the agreement score provides a compact measure of consistency, the rationales help expose why different language-specific judgments agree or conflict. In particular, they provide useful signals about missing constraints, vague definitions, or underspecified decision boundaries in the current criterion.

\paragraph{Iterative Criterion Refinement}
We iteratively refine the judging process using the multilingual inconsistency collected at each iteration. At iteration $t+1$, the judge predicts by conditioning on: i) the original evaluation context $C$ and ii) the memory that stores accumulated feedback from previous iterations $M^{(t)}$. This feedback history consists of the cross-lingual agreement scores, together with the corresponding judgments and rationales obtained independently from each language variant. By augmenting the original context with such historical feedback, the judge can make a more informed prediction for the same sample in the next iteration. Formally, the refinement process is defined as:
\begin{equation}
p^{(t+1)} \leftarrow {\rm{\tt LLM}}(C \otimes M^{(t)} \otimes x),
\end{equation}
We terminate the refinement once the agreement across multilingual variants exceeds a threshold $\tau$. To prevent excessive refinement, we also impose a maximum number of iterations, denoted by $s$.

\section{Experiments Setup}
\subsection{Data}
\paragraph{Dataset} We evaluate on five multilingual benchmarks: XCOPA \citep{ponti-etal-2020-xcopa} for causal reasoning, Wino-X \citep{emelin-sennrich-2021-wino} for commonsense reasoning, XStoryCloze \citep{lin-etal-2022-shot} for story understanding, Belebele \citep{bandarkar-etal-2024-belebele} for reading comprehension, and MMLU-ProX \citep{xuan-etal-2025-mmlu} for knowledge-intensive question answering. These datasets span diverse tasks and languages, enabling comprehensive evaluation. They provide aligned cross-lingual samples with explicit correct and incorrect options, which allows us to systematically construct judgment instances for LLM-as-a-judge evaluation.

\paragraph{Data Construction}
To construct reliable binary judgment data, we reformulate each example into a binary classification instance. Since existing benchmarks are often dominated by positive labels, directly using the original data may introduce label bias into the judge. To mitigate this issue, we explicitly construct balanced positive and negative instances. For each example, we randomly pair the input with either its correct answer or an incorrect answer with equal probability, and assign the corresponding binary judgment label. This procedure ensures a uniform distribution over positive and negative cases. For example, given the Wino-X instance ``The bag ripped when the boy placed a toy in it because it was too flimsy.'' (input), with candidate answers ``the bag'' (correct) and ``the toy'' (incorrect), we construct the following binary judgment instances and randomly sample one:
\begin{center}
\begin{minipage}{0.9\linewidth}
\small
Sentence: \{input\}; Answer: \{the bag\} $\rightarrow$ correct \\
Sentence: \{input\}; Answer: \{the toy\} $\rightarrow$ incorrect
\end{minipage}
\end{center}
This balanced construction prevents the judge from relying on label priors and instead encourages it to distinguish between correct and incorrect candidate answers based on semantic reasoning. Data statistics is provided in Appendix \ref{app:data_stat}.

\subsection{Baselines}
To evaluate the effectiveness of SEMJ, we compare against several widely used judging strategies that enhance the reasoning process in different ways:

\noindent\textbf{Base Judge} directly applies the original model to produce judgments without any refinement.
    
\noindent\textbf{Multilingual Vote} extends the input into multiple language variants and aggregates their predictions via majority voting \citep{qin-etal-2023-cross,wang-etal-2025-calm}.
    
\noindent\textbf{Monolingual Vote} samples multiple independent judgments from the same input and aggregates them via majority voting \citep{wang2023selfconsistency}.
    
\noindent\textbf{Monolingual Reflection} iteratively refines the judgment process using only the original-language input, without leveraging multilingual consistency signals \citep{madaan2023selfrefine}.

\begin{table*}
    \centering
    \resizebox{\textwidth}{!}{
    \renewcommand{\arraystretch}{1.2}
    \begin{tabular}{lcccccccccccc} \toprule
         \multirow{3}{*}{Model}  & \multicolumn{2}{c}{XCOPA}  & \multicolumn{2}{c}{Wino-X} & \multicolumn{2}{c}{Wino-X-Mt}  & \multicolumn{2}{c}{XStoryCloze}  &  \multicolumn{2}{c}{Belebele} & \multicolumn{2}{c}{MMLU-ProX} \\ \cmidrule(r){2-3} \cmidrule(r){4-5} \cmidrule(r){6-7} \cmidrule(r){8-9} \cmidrule(r){10-11} \cmidrule(r){12-13}
         & Acc &FK & Acc &FK & Acc &FK & Acc &FK & Acc &FK & Acc &FK \\ \midrule
         Base Judge &58.64 &0.4049 &66.96 &0.2152 &48.97 &0.2563  &74.49  &0.2387  &62.53  &0.1975 &56.18  &0.5534 \\
         Multilingual Vote  &49.00 &- &50.44 &-  &50.15 &-  &75.28  &-  &63.70  &- &57.14  & - \\
         Monolingual Vote  &59.00 &0.4349 &67.86 &0.2268 &48.08 &0.2319  &74.42  &0.2379  &62.54  & 0.1914&56.11  &0.5548  \\
         Monolingual Ref &60.69 &0.4211 &68.75 &0.2935 &48.97 &0.2209  &75.56  &0.2100  &63.46  &0.1965 &58.49  &0.4951  \\
         SEMJ (ours)  &\textbf{63.36} &\textbf{0.4470} &\textbf{72.92} &\textbf{0.3922} &\textbf{50.44} &\textbf{0.3156} &\textbf{78.84} &\textbf{0.3207} &\textbf{63.92} &\textbf{0.2175} &\textbf{60.27}  &\textbf{0.5738}  \\ 
         \bottomrule
    \end{tabular}}
    \caption{Performance of multilingual LLM-as-a-Judge across six datasets. Fleiss's Kappa (FK) is calculated for both settings to measure judgment consistency across parallel data.}
    \label{tab:main_rst}
\end{table*}

\subsection{Implementation Details}
We conduct our main experiments using the open-source model Qwen-2.5-7B-Instruct \citep{qwen2.5}, given its strong overall performance. To further evaluate the generalizability of SEMJ, we also experiment with additional base models, including Llama-3-8B-Instruct \citep{grattafiori2024llama}, the multilingual-pretrained BLOOMZ-7B \citep{muennighoff-etal-2023-crosslingual}, as well as superior open-source models such as GPT-4.1 \citep{openai2025gpt41} and Claude-4.5-Sonnet \citep{anthropic2025claude-opus-4-5}.


For multilingual sampling in SEMJ, the judge model itself generates the multilingual variants.\footnote{The detailed translation prompts are provided in Appendix~\ref{app:translation}.} We set the sampling number $K$ to 4 by randomly selecting judgments from other languages, as studied in \S\ref{exp:sampling}. For self-evolution, we set the early termination threshold ($\tau$) to 0.8 and the maximum number of evolution iterations $s$ to 3. To mitigate randomness, we run all experiments five times and report the average results. Additional implementation and hyperparameter selection are provided in Appendix \ref{app:hyperparameter}, and prompts are in Appendix \ref{exp:prompts}.

\subsection{Evaluation Metrics}
 

Following prior work on multilingual LLM-as-a-Judge evaluation \citep{fu-liu-2025-reliable}, we adopt Fleiss’ Kappa (FK) to measure cross-lingual judgment consistency, where higher FK indicates more stable behavior across language variants. However, consistency does not necessarily imply correctness, as a judge may remain consistent while being systematically wrong. Therefore, we additionally report accuracy to measure whether the predicted labels match the ground-truth answers.


\section{Results and Analysis}
\subsection{Main Results}
\paragraph{Overall Results} Table \ref{tab:main_rst} presents the performance of multilingual LLM-as-a-Judge models across various benchmarks. Compared to the Base Judge, our proposed SEMJ consistently achieves higher consistency (FK) on nearly all tasks. For instance, the score improves from 0.2152 to 0.3922 ($\Delta$ 0.177) on Wino-X, and from 0.2380 to 0.3207 ($\Delta$ 0.0827) on XStoryCloze. These results align with our motivation of encouraging the judge to maintain consistent evaluations across semantically aligned multilingual inputs. To rule out the possibility of trivial consistency (e.g., collapsing to identical predictions), we further evaluate judgment accuracy. Table \ref{tab:main_rst} shows accuracy improves from 66.96 to 72.92 ($\Delta$ 5.96) on Wino-X, and from 74.49 to 78.84 ($\Delta$ 4.35) on XStoryCloze, with similar trends observed across other datasets. Overall, these results demonstrate that SEMJ effectively enhances judge quality and multilingual consistency.

We further observe that \textit{Multilingual Vote} performs unstably, and even substantially underperforms the Base Judge on XCOPA and Wino-X. This suggests that multilingual judgments indeed contain useful diagnostic signals, but are often too noisy to be directly aggregated through naive voting. Meanwhile, \textit{Monolingual Vote} performs similarly to the Base Judge across datasets, suggesting that repeated sampling within a single language fails to sufficiently expose complementary judge information due to limited decoding diversity. In contrast, \textit{Monolingual Reflection} consistently improves judgment accuracy over the Base Judge, increasing performance from 66.96 to 68.75 ($\Delta$ 1.79) on Wino-X and from 74.49 to 75.56 ($\Delta$ 1.07) on XStoryCloze. These gains suggest that reassess and refine its own reasoning process produce more reliable judgments. However, Monolingual Reflection still underperforms SEMJ, indicating that SEMJ benefits not only from iterative refinement, but also from complementary signals introduced by multilingual inconsistency.

\begin{table}
    \centering
    \resizebox{\columnwidth}{!}{
    \begin{tabular}{llcccc} \toprule
         &\multirow{3}{*}{Model}  & \multicolumn{2}{c}{XCOPA}  & \multicolumn{2}{c}{XStoryCloze} \\ \cmidrule(r){3-4} \cmidrule(r){5-6} 
         && Acc &FK & Acc &FK  \\ \midrule
         \multirow{2}{*}{LLama-3} &Base &58.55 &0.1238 &71.60 &0.2327   \\
         &SEMJ &60.60 &0.2344 &73.32 &0.2678 \\ \hline
         \multirow{2}{*}{Bloomz} &Base &53.71 &0.1349 &68.37 &0.1915   \\
         &SEMJ &56.43 &0.1768 &70.44 &0.2136 \\ \hline
         \multirow{2}{*}{GPT-4.1} &Base & 82.82 &0.1771 &94.60 &0.2879  \\
         &SEMJ &85.36 &0.2651 &96.93  &0.3619 \\ \hline
         \multirow{2}{*}{Claude-4.5-Sonnet} &Base &85.45 &0.1779 &92.48 &0.4377   \\
         &SEMJ &88.55 &0.2704 &95.20 &0.5367 \\
         \bottomrule
    \end{tabular}}
    \caption{Performance of SEMJ and original Base Judge across various judge backbones from different LLM families.}
    \label{tab:generalization}
\end{table}


\paragraph{Generalization}
To further evaluate the generalizability of SEMJ, we conduct experiments on multiple base models, including open-source models (Llama-3 and Bloomz) and proprietary models (GPT-4.1 and Claude-4.5). As shown in Table~\ref{tab:generalization}, SEMJ consistently improves both judgment accuracy and cross-lingual consistency across different model families. Notably, even for strong proprietary models such as GPT-4.1, SEMJ still achieves clear gains on both XCOPA and XStoryCloze, suggesting that the proposed strategy is not restricted to a specific architecture or model scale. 

\begin{figure}
    \centering
    \includegraphics[width=\linewidth]{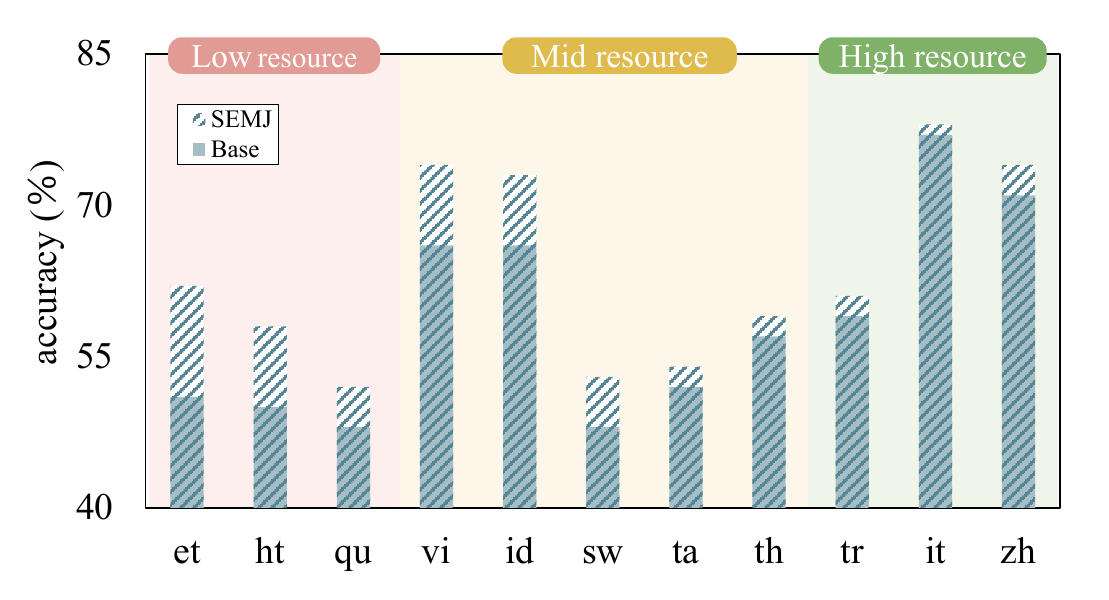}
    \caption{Performance comparison of Base and SEMJ on the XCOPA dataset across languages of different resource levels.}
    \label{fig:lang_diff}
\end{figure}

\paragraph{Language Results}
We evaluate the effectiveness of SEMJ on the XCOPA dataset across 11 languages spanning different resource levels.\footnote{Resource levels are defined as a coarse-grained adaptation of \citep{joshi-etal-2020-state}, reflecting relative availability of language resources.} As shown in Figure \ref{fig:lang_diff}, we group languages into low-resource (et, ht, qu), mid-resource (vi, id, sw, ta, th, tr), and high-resource (it, zh) categories to analyze performance under varying data availability conditions. Overall, SEMJ consistently outperforms the Base model across all languages, demonstrating its robustness in multilingual settings. Notably, the improvements are more pronounced in low- and mid-resource languages, suggesting that SEMJ is particularly effective in mitigating performance degradation under data scarcity. Although the gains in high-resource languages are relatively smaller, they remain consistent, suggesting that SEMJ’s benefits are not merely due to compensating for limited resources, but also arise from leveraging complementary perspectives across languages.

\subsection{Analysis}
We further explore \textit{why does multilingual self-evolution improve multilingual LLM judges?}

\paragraph{Does Inconsistency Trigger Reconsideration?}
We first examine how cross-lingual inconsistency influences the multilingual judging process in SEMJ. To isolate this effect from iterative refinement, we focus on the transition between the base judge and the first evolution step. We bucket examples by their consistency scores and measure two metrics after evolution: i) \textit{flip rate}, the fraction of judgments that change after evolution; and ii) \textit{repair rate}, the fraction of incorrect judgments corrected after evolution. Both metrics are normalized within each bucket. Table \ref{tab:flip_rate} shows a clear trend.

We observe that both flip rate and repair rate are substantially lower under high-consistency settings ($cr \in [0.75,1]$), while lower-consistency examples exhibit nearly three times higher rates. This suggests that \textit{multilingual inconsistency significantly increases the likelihood that the judge revisits its prior decision}. Moreover, the comparable flip and repair rates across the 0.5--0.75 consistency buckets indicate that the presence of conflict, rather than its exact degree, is the primary trigger for re-evaluation. However, repair rate remains consistently lower than flip rate, suggesting that inconsistency mainly promotes reconsideration rather than guaranteeing more accurate judgments.

\begin{table}[t]
\centering
\small
\begin{tabular}{lcc}
\toprule
Consistency Rate & Flip Rate & Repair Rate \\
\midrule
$cr=[1, 0.75)$    & 5.72  & 2.95  \\
$cr=[0.75, 0.50)$ & 18.14 & 11.93 \\
$cr=[0.50, 0.25)$ & 17.57 & 13.52 \\
\bottomrule
\end{tabular}
\caption{Flip and repair rates under different cross-language consistency rate. }
\label{tab:flip_rate}
\end{table}


\paragraph{What Determine Successful Repair?}
After inconsistency triggers reconsideration, we further investigate what distinguishes successful repair cases. We analyze repaired samples using Sparse Autoencoders (SAE) \citep{cunningham2024sparse}, which maps a residual-stream activation into a sparse latent space and reconstructs the original activation from only a small set of active features, making these latent dimensions more interpretable. For each repaired example, we run the same judge model twice. Once with the original prompt and once with the evolved prompt, while keeping the evaluated instance fixed. We extract residual-stream activations at answer-relevant positions and project them through a pretrained SAE\footnote{\url{https://huggingface.co/andyrdt/saes-qwen2.5-7b-instruct}}. For each latent feature $f$, we compute the activation difference between the evolved and original prompts:
\begin{equation}
\Delta a_f(x) = a_f(p_{\mathrm{evo}}, x) - a_f(p_{\mathrm{orig}}, x),
\end{equation}
where $x$ denotes the evaluated example, $p_{\mathrm{orig}}$ is the original judge prompt, and $p_{\mathrm{evo}}$ is the evolved prompt. We then focus on the top activated latent dimensions with the largest activation increases. To interpret these dimensions, we collect examples with strong activation increases and contrast them with hard negative cases where evolution fails to repair the judgment and the same latent remains weakly activated. Based on positive and negative examples, we prompt GPT-4.1 \citep{openai2025gpt41} to generate interpretations for each latent feature.

\begin{table}[t]
\centering
\small
\begin{tabular}{p{0.18\linewidth} p{0.70\linewidth}}
\toprule
\textbf{Latent} & \textbf{Explanation} \\
\midrule
f64514 & \textit{Corrective evidence}. The history contains premise-grounded evidence relevant to the gold judgment. \\ \midrule
f18444 & \textit{Causal rationale}.  The added rationale gives a concrete causal mechanism supporting revision. \\ \midrule
f62970 & \textit{Prior challenge}. The new rationale directly contradicts or overturns the prior verdict. \\ \midrule
f36698 & \textit{Evidence weighing}.  Correct-side rationales show stronger rubric-aligned evidence than alternatives. \\
\bottomrule
\end{tabular}
\caption{Representative SAE latents associated with SEMJ's corrective evidence mechanism analized on XCOPA.}
\label{tab:sae_latents}
\end{table}

Table \ref{tab:sae_latents} shows four representative activated dimensions along with their corresponding explanations. We observe that these activated dimensions correspond to fine-grained reasoning signals, such as concrete evidence, causal rationale, prior challenges, and evidence weighting.\footnote{Appendix~\ref{app:sae_examples} provides concrete examples for each representative SAE latent.} These findings suggest a two-stage mechanism. Cross-lingual inconsistency first serves as a reconsideration trigger, reducing the model’s confidence in its initial verdict. Subsequently, the model requires usable evidence to determine how to revise its judgment. The SAE results indicate that evolved prompts are most effective when the historical context contains corrective rationales that are specific and sufficiently strong to override the prior decision.

\begin{table}
    \centering
    \resizebox{\columnwidth}{!}{
    \begin{tabular}{cccccccc} \toprule
         & $K$=1  &$K$=2 & $K$=3  &$K$=4  &$K$=5 &$K$=8 &$K$=10  \\ \midrule
        Acc &60.74  &61.06  &61.92  &63.36  &63.94 &64.18 &64.30 \\
        $\Delta$ &-  &+0.32  &+1.44  &+2.30  &+0.58 &+0.24 &+0.12 \\ \bottomrule
    \end{tabular}}
    \caption{Effect of the number of parallel translated aligned samples $K$ on judge accuracy. $\Delta$ indicates the accuracy gain over the previous $K$ setting.}
    \label{tab:parallelk}
\end{table}

\subsection{What Matters for SEMJ?}
In this section, we analyze how design choices in SEMJ affect the performance of our method.

\paragraph{Impact of Language Composition}
\label{exp:sampling}
We first examine how the number of parallel translated language pairs influences performance.
Specifically, we vary the sampling size $K$ and compare the resulting judgment performance. Table \ref{tab:parallelk} shows that increasing the number of parallel translated samples $k$ consistently improves judgment accuracy. Meanwhile, the gains ($\Delta$) exhibit a generally monotonic trend, with the largest improvement observed at $K=4$. This suggests that while additional languages provide useful complementary signals for the multilingual judge, their benefit is bounded. Beyond a certain point, introducing more languages becomes less effective, likely because their largely overlapping semantic information.

\begin{table}
    \centering
    \resizebox{\columnwidth}{!}{
    \begin{tabular}{cccccc} \toprule
         & random  &high lang & low lang  &same family  &diff family \\ \midrule
        Acc &63.36  &61.95 &63.15  &61.97  &62.06  \\
        FK &0.4470  &0.4385  &0.4343  &0.4475  &0.4577 \\ \bottomrule
    \end{tabular}}
    \caption{Effect of different language selection strategies for choosing the $K$ translated languages.}
    \label{tab:kfamily}
\end{table}

We further study whether the gains of SEMJ depend on specific language selection strategies. We compare random sampling with several heuristic-based strategies, including selecting from high-resource languages, low-resource languages, same-family languages, and different-family languages. As shown in Table~\ref{tab:kfamily}, random sampling achieves the best overall performance. This suggests that the benefit of SEMJ mainly comes from diversity of evaluation perspectives rather than carefully designed language compositions.

\paragraph{Effect of Iterative Evolution Rounds}

\begin{table}
    \centering
    \resizebox{0.9\columnwidth}{!}{
    \begin{tabular}{cccccc} \toprule
         & $s$=1  &$s$=2 & $s$=3  &$s$=4  &$s$=5\\ \midrule
        Acc &62.64  &62.73  &63.36  & 62.87 &62.20  \\
        FK &0.4409  &0.4474  &0.4470  &0.4595  &0.4606   \\ \bottomrule
    \end{tabular}}
    \caption{Effect of the number of iterative self-evolution rounds $s$ on judge accuracy and consistency. }
    \label{tab:s}
\end{table}

We explore the effect of iterative self-evolution rounds in SEMJ by varying the number of evolution steps $s$. Table~\ref{tab:s} shows the accuracy exhibits a monotonic trend as $s$ increases, reaching its peak at $s=3$. This suggests that iterative evolution can progressively refine judge decisions and improve performance to a certain extent. In contrast, the consistency score continues to increase throughout all evolution rounds. The difference between accuracy and consistency reveals a mild overfitting phenomenon, where excessive evolution enforces cross-lingual consistency hence gradually suppressing beneficial diversity and complementary disagreement signals across languages. Consequently, the judge becomes more self-consistent but less accurate. Therefore, we select $s=3$, which achieves the best trade-off between multilingual consistency and judge accuracy.

\section{Conclusion}
In this work, we show that multilingual judge inconcsistency serves as a complementary signal instead of noise. Building on this insight, we propose an inconsistency-driven self-evolving multilingual judge framework, which consistently improves performance across multiple benchmarks and model backbones. 
We hope this work inspires future research to rethink multilingual inconsistency as a constructive signal for developing more reliable language aware evaluation systems.

\section{Limitations}
The proposed SEMJ method introduces additional inference cost due to the iterative self-evolution process and multilingual sampling, which requires multiple rounds of model inference across different languages. This increases computational overhead compared to standard single-pass judgment models. However, the evolved data can be reused to distill the capability into a more efficient model, for example through supervised fine-tuning \citep{wang-etal-2023-self-instruct} or preference optimization (e.g., DPO \citep{yuan2024selfrewarding}), by constructing training or preference pairs from the evolution trajectories. In this way, the inference-time cost can be largely eliminated while retaining the benefits of SEMJ during deployment.

\bibliography{custom,anthology-1}

\clearpage

\appendix
\begin{strip}
\centering
\footnotesize
\renewcommand{\arraystretch}{1.2}
\resizebox{\textwidth}{!}{
\begin{tabular}{@{}llllcc@{}}
\toprule
Dataset & Task & Answer Type & Languages & Train & Test \\
\midrule
\makecell[l]{\textbf{XCOPA} \\ \citet{ponti-etal-2020-xcopa}}
& \makecell[l]{Causal Commonsense \\ Reasoning}
& Binary Choice
& \makecell[l]{Estonian, Haitian Creole, Indonesian, Italian, Quechua, \\ Swahili, Tamil, Thai, Turkish, Vietnamese, Chinese}
& 400 & 200 \\
\midrule

\makecell[l]{\textbf{Wino-X} \\ \citet{emelin-sennrich-2021-wino}}
& Coreference Resolution
& Binary Choice
& German, French, Russian, English
& - & 84 \\
\midrule

\makecell[l]{\textbf{Wino-X-Mt} \\ \citet{emelin-sennrich-2021-wino}}
& Machine Translation
& Binary Choice
& German, French, Russian
& - & 113 \\
\midrule

\makecell[l]{\textbf{XStoryCloze} \\ \citet{lin-etal-2022-shot}}
& Story understanding
& Binary Choice
& \makecell[l]{English, Russian, Chinese (Simplified), Spanish (Latin America), Arabic, Hindi, \\ Indonesian, Telugu, Swahili, Basque, Burmese}
& 1510 & 360 \\
\midrule

\makecell[l]{\textbf{Belebele} \\ \citet{bandarkar-etal-2024-belebele}}
& \makecell[l]{Machine Reading \\ Comprehension}
& Multiple Choice
& 122 language variants, but 115 distinct languages (ignoring scripts)
& 765 & 135 \\
\midrule

\makecell[l]{\textbf{MMLU-ProX} \\ \citet{xuan-etal-2025-mmlu}}
& Question Answering
& Multiple Choice
& 29 typologically diverse languages
& 11759 & 1764 \\
\bottomrule
\end{tabular}
}
\captionof{table}{Datasets for multilingual LLM-as-a-Judge evaluation, all involving parallel data across provided languages. \textit{Train} and \textit{Test} indicate the number of training and testing samples per language. For all experiments, we report results on the test split.}
\label{tab:data_stat}
\end{strip}

\section{Data Statistics}
\label{app:data_stat}
Table~\ref{tab:data_stat} summarizes the statistics of the datasets used in our experiments. Our evaluation covers different multilingual benchmarks with diverse task formats and language coverage. These datasets provide a comprehensive assessment of multilingual LLM-as-a-judge. 

For each benchmark, we split the data into training and test sets. All main results are reported on the test sets to ensure a clean and unbiased comparison, while the training sets are used solely for hyperparameter selection. These training splits are also useful for future work, such as supervised judge training.

\section{Implementation Details}
\subsection{Hyperparameter Selection}
\label{app:hyperparameter}
Our method introduces a threshold hyperparameter $\tau$ to control the termination of the self-evolution process. We determine the optimal value of $\tau$ through hyperparameter tuning on the training split of XCOPA with ten aligned multilingual samples by evaluating a range of candidate thresholds. Smaller values of $\tau$ tend to terminate the evolution prematurely, limiting the model’s ability to sufficiently refine its judgments, whereas larger values permit excessive evolution iterations, which may lead to over-evolution and diminishing returns. Table~\ref{app:tau} presents the results. We observe that thresholds around $\tau = 0.8$ consistently achieve the best overall accuracy. Therefore, we fix it for all experiments in this work.

\subsection{Prompts}
\label{exp:prompts}
\paragraph{Judge Prompt} Our prompt template is based on the multilingual judge design of \citet{fu-liu-2025-reliable}, with additional components introduced to support SEMJ’s self-evolution process, including cross-lingual historical judgments and consistency feedback (\S \ref{judging principle}). Table~\ref{app:prompt} presents the prompt templates used in our multilingual LLM-as-a-Judge framework. Table~\ref{app:rubric} details the dataset specific formulations of the \texttt{<rubric>} and \texttt{<input>} fields across different benchmarks.

\begin{table}
    \centering
    \resizebox{0.7\columnwidth}{!}{
    \begin{tabular}{ccccc} \toprule
         & $\tau$ = 0.6 &$\tau$ = 0.7 &$\tau$ = 0.8 &$\tau$ = 0.9  \\ \midrule
        Acc   &61.47  &62.36  &64.30  & 63.11 \\
        $\Delta$ &0.4129  &0.4435  &0.4582  &0.4639  \\ \bottomrule
    \end{tabular}}
    \caption{Effect of the number of parallel translated aligned samples $k$ on judge accuracy. $\Delta$ indicates the accuracy gain over the previous $k$ setting.}
    \label{app:tau}
\end{table}

\begin{table}[!t]
    \centering
    \small
    \begin{tabular}{p{7.3cm}} \toprule
    \cellcolor{C5}<role definition>\\
    \cellcolor{C5}You are an AI assistant whose purpose is to evaluate the correctness of answers to questions in \{language\_tag\}.    \\ 
    \cellcolor{C2}<rubric> \\
    \cellcolor{C2}Your evaluation should consider correctness and helpfulness. Do not allow the length of the answer to influence your evaluation. Be as objective as possible. \\
    \cellcolor{C4}<evolve history>\\
    \cellcolor{C4}History judge from previous iteration: \\
    \cellcolor{C4}- round index: \{iter\_num\} \\
    \cellcolor{C4}- crosslingual consistency rate: \{consistency\_rate\}  \\
    \cellcolor{C4}- judgments and rationales: \{lang\}: \{judge\}\{rationale\} \\
    \cellcolor{C4}Please reconsider your judgment by incorporating cross-lingual historical judgments as additional diagnostic signals for the current evaluation. \\
    \cellcolor{C1}<output specification>\\
    \cellcolor{C1}Respond with a single JSON object and nothing else, with exactly these keys: `correct' (boolean), `reason' (string). \\
    \cellcolor{C3}<input> \\ \bottomrule
    \end{tabular}
    \caption{Prompt template for multilingual LLM-as-a-Judge. The <rubric> and <input> fields are instantiated differently for each dataset. Detailed prompt specifications are provided in Table~\ref{app:rubric}.}
    \label{app:prompt}
\end{table}

\begin{table*}
    \centering
    \small
    \begin{tabular}{lp{13.3cm}} \toprule
    Dataset &Prompt  \\ \midrule
    \multirow{4}{*}{XCOPA} & \textit{rubric}: You are given a premise sentence, a question type indicating either a cause or an effect, and a proposed anwer. Your task is to determine whether the proposed answer correctly identifies the most plausible cause or effect of the premise. \\ \cmidrule(r){2-2}
    &\textit{input}: Premise: \{input\}; Question Type: \{rel\}; Answer: \{output\} \\ \midrule
    \multirow{4}{*}{Wino-X} & \textit{rubric}: You are given a sentence containing a blank and a proposed answer that fills the blank. Your task is to determine whether the filled-in answer correctly completes the sentence. \\ \cmidrule(r){2-2}
    &\textit{input}: Sentence: \{input\}; Answer: \{output\} \\ \midrule
    \multirow{4}{*}{Wino-X-Mt} & \textit{rubric}: You are given a source sentence and a proposed translation. Your task is to determine whether the proposed translation accurately translates the source sentence. \\ \cmidrule(r){2-2}
    &\textit{input}: Source: \{input\}; Target: \{output\} \\ 
    \midrule
    \multirow{4}{*}{XStoryCloze} & \textit{rubric}: You are given a story context, and a potential ending. Your task is to determine whether the ending completes the story. \\ \cmidrule(r){2-2}
    &\textit{input}: Context: \{input\}; Ending: \{output\}\\ \midrule
    \multirow{4}{*}{Belebele} & \textit{rubric}: You are given a context, a question, and an answer. Your task it to determine whether the generated answer is correct according to the provided context.\\ \cmidrule(r){2-2}
    &\textit{input}: Context and Question: \{input\}; Answer: \{output\}\\ \midrule
    \multirow{4}{*}{MMLU-Prox} & \textit{rubric}: You are given a question and an answer. Your task it to determine whether the generated answer is correct.\\ \cmidrule(r){2-2}
    &\textit{input}: Question: \{input\}; Answer: \{output\}\\  \bottomrule
    \end{tabular}
    \caption{Rubric and input prompts for different benchmarks.}
    \label{app:rubric}
\end{table*}

\paragraph{Translation Prompt}
\label{app:translation}
We use the prompt shown in Table\ref{app:translate_prompt} to perform self-translation for all language pairs. The judge model is instructed to translate each input into the target language while strictly preserving semantic meaning, without adding any extra information or explanations. The prompt is kept fixed across all experiments to ensure consistency.

\section{Complmentary in Multilinual Sampling}
\label{app:complimentary}

\begin{table}[!t]
    \centering
    \small
    \begin{tabular}{p{7.3cm}} \toprule
    You are a multilingual translator. Translate the following text into \{language\_tag\} while preserving its original meaning as faithfully as possible.\\
    Do not change the semantic content. Keep the translation natural and fluent in the target language. Preserve names, labels, and formatting.    \\ 
    Input: \{input\} Output: Return only the translated text in \{language\_tag\} \\ \bottomrule
    \end{tabular}
    \caption{Translation prompt used for constructing semantically aligned pairs.}
    \label{app:translate_prompt}
\end{table}

To verify whether the observed complementarity in multilingual $k$-oracle generalizes beyond XCOPA, we further conduct experiments on two additional benchmarks: \textit{MMlu-Prox} and \textit{Belebele}, covering differenct size of language numebrs.

\begin{figure}
    \centering
    \includegraphics[width=0.8\linewidth]{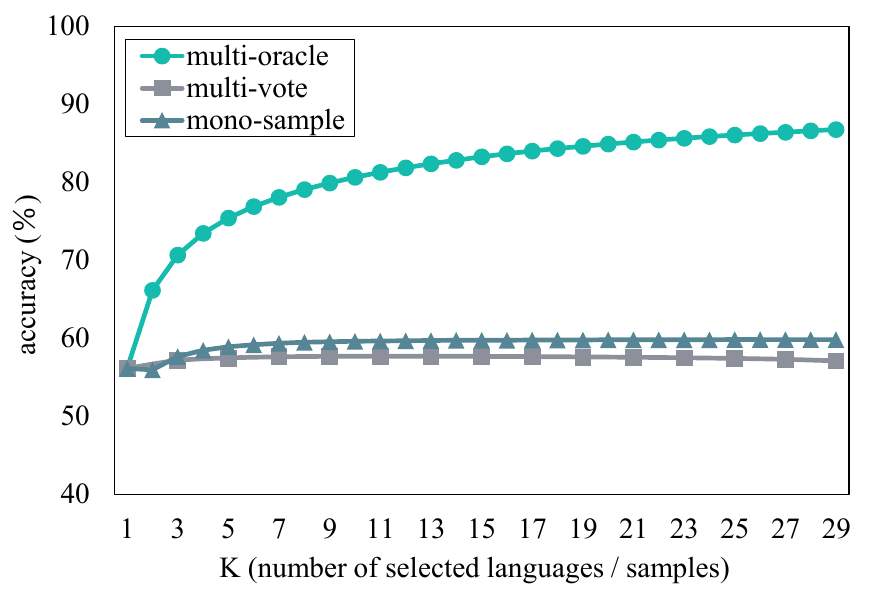}
    \caption{Performance of complementarity in multilingual sampling on MMLU. We compare multilingual oracle (multi-oracle), multilingual majority voting (multi-vote), and monolingual $K$-sampling (mono-sample) under different values of $K$, i.e, the number of selected languages or samples. }
    \label{app:compli_mmlu}
\end{figure}

\begin{figure}
    \centering
    \includegraphics[width=0.8\linewidth]{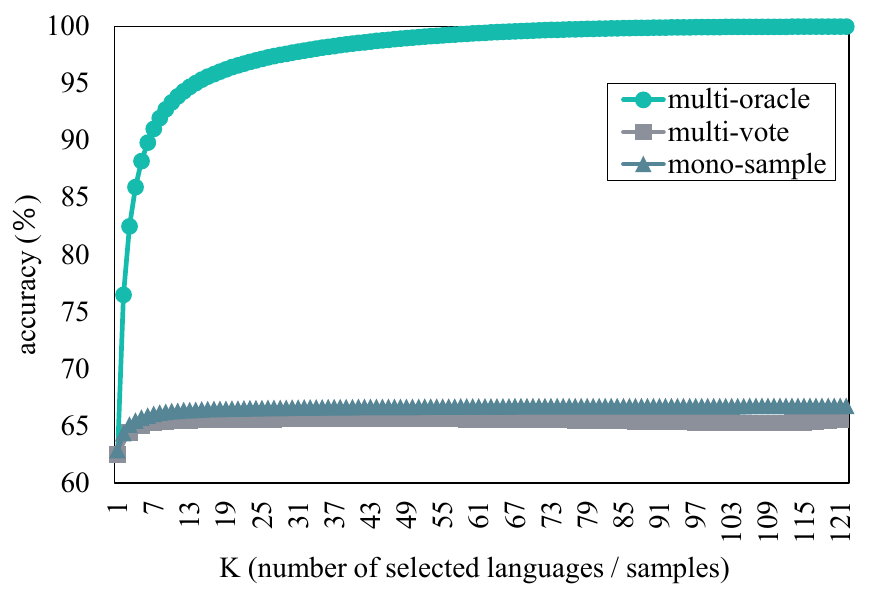}
    \caption{Performance of complementarity in multilingual sampling on Belebele. We compare multilingual oracle (multi-oracle), multilingual majority voting (multi-vote), and monolingual $K$-sampling (mono-sample) under different values of $K$, i.e, the number of selected languages or samples. }
    \label{app:compli_belebele}
\end{figure}

As shown in Figure~\ref{app:compli_mmlu} and Figure~\ref{app:compli_belebele}, we observe a consistent trend across both datasets, the multilingual $k$-oracle accuracy steadily increases as the number of languages $k$ grows. This indicates that the gain from multilingual aggregation is not specific to a single dataset or task type, but rather a general phenomenon across different reasoning and commonsense understanding tasks. We further compare against the same two baselines used in the main analysis, i.e., multilingual majority voting (multi-vote) and mono-lingual $k$-sample. In both datasets, the multilingual $k$-oracle consistently outperforms these baselines across all values of $k$, reinforcing that the improvement cannot be explained by simple ensembling effects or increased sampling budget.

Overall, these additional results corroborate the findings in Section~\ref{sec:complementary}, suggesting that multilingual inconsistency systematically encodes complementary correctness signals across languages rather than being purely stochastic noise.

\section{SAE Latent Examples}
\label{app:sae_examples}

Table~\ref{app:sae_example1} and Table\ref{app:sae_example2} show representative examples for the selected SAE latents. These examples illustrate how different latents capture premise-grounded evidence, causal rationales, prior-verdict challenges, and evidence weighing in the evolved prompt.

\begin{table}[h]
    \centering
    \small
    \begin{tabular}{p{7cm}} \toprule
    \textit{\textbf{f64514}}: \textit{Corrective evidence}. The history contains premise-grounded evidence relevant to the gold judgment.    \\ 
    \textbf{input(zh)}: Premise: {\pchn{秘书搁置了来电人。}}(The secretary put the caller on hold.); Question Type: effect; Answer: {\pchn{来电人在线上等着。}}(The caller waited on the line.) \textbf{ground truth}: True \\ \midrule
    \cellcolor{C2}round 0: \\
    \textbf{judge(zh)}: False; The secretary transferred the call to someone else so the caller didn't have to wait any longer. \\
    \textbf{judge(tr)}: True; When the secretary puts the caller on hold, it results in the caller waiting, which is a plausible and direct consequence of the action described in the premise.\\
    \cellcolor{C2}round 1: \\
    \textbf{judge(zh)}: True; Having the caller waiting online is a direct consequence of ignoring the caller.\\
    \bottomrule
    \end{tabular}
    \caption{Examples of representative SAE latents associated with SEMJ’s corrective evidence mechanism on XCOPA.}
    \label{app:sae_example1}
\end{table}

\begin{table}[t]
    \centering
    \small
    \begin{tabular}{p{7cm}} \toprule
    \textit{\textbf{f18444}}: \textit{Causal rationale}. The added rationale gives a concrete causal mechanism supporting revision.    \\ 
    \textbf{input(vi)}: Premise: \foreignlanguage{vietnamese}{Ngọn lửa trên cây nến đã tắt.} (The flame on the candle has gone out.); Question Type: effect; Answer: \foreignlanguage{vietnamese}{Tôi thổi vào bấc.} (I blew on the wick.) \textbf{ground truth}: True \\ \midrule
    \cellcolor{C2}round 0: \\
    \textbf{judge(vi)}: False; The premise states that the candle flame has already gone out, so blowing out the candle would be a cause, not an effect. \\
    \textbf{judge(it)}: True; Blowing on the flame can extinguish it by depriving it of oxygen or directly cooling it.\\
    \cellcolor{C2}round 1: \\
    \textbf{judge(vi)}: True; Based on the information provided, ``I blew out the wick" is a reasonable cause-and-effect relationship, because blowing out the wick can directly cause the flame to go out.\\ \midrule\midrule
    \textit{\textbf{f62970}}: \textit{Prior challenge}. The new rationale directly contradicts or overturns the prior verdict. \\ 
    \textbf{input(tr)}: Premise: \foreignlanguage{turkish}{At tökezledi.}(The horse stumbled.); Question Type: cause; Answer: \foreignlanguage{turkish}{Atı bir sinek ısırdı.}(A fly bit the horse.) \textbf{ground truth}: True \\ \midrule
    \cellcolor{C2}round 0: \\
    \textbf{judge(tr)}: False; The answer is incorrect because “a fly bit the horse” does not logically follow as an effect of the horse kicking; it describes a separate event rather than a consequence of the premise.\\
    \textbf{judge(id)}: True; Being bitten by flies can cause discomfort or pain to a horse, which can lead to it stumbling.\\
    \cellcolor{C2}round 1: \\
    \textbf{judge(tr)}: True; The answer is correct because a fly bite can irritate or hurt the horse, making it plausible that the horse would suddenly kick. \\ \midrule\midrule
    \textbf{\textit{f36698}}: \textit{Evidence weighing}.  Correct-side rationales show stronger rubric-aligned evidence than alternatives. \\ 
    \textbf{input(sw)}: Premise: Mwanamke alimwomba mwanaume aondoke. (The woman asked the man to leave.); Question Type: cause; Answer: Alimshukuru. (He thanked her.) \textbf{ground truth}: False \\ \midrule
    \cellcolor{C2}round 0: \\
    \textbf{judge(sw)}: False; The answer is incorrect because ``he thanked her'' is not a plausible cause of the woman asking him to leave. \\
    \textbf{judge(zh)}: False; Thanking her is unlikely to be the reason the woman asked him to leave. A more plausible cause is that he did something inappropriate or made her uncomfortable.\\
    \cellcolor{C2}round 1: \\
    \textbf{judge(sw)}: False; The answer remains incorrect: ``thanking her'' is better viewed as an unrelated social response, while a more plausible cause would be that the man did something inappropriate or made her uncomfortable. \\
    \bottomrule
    \end{tabular}
    \caption{Examples of representative SAE latents associated with SEMJ’s corrective evidence mechanism on XCOPA.}
    \label{app:sae_example2}
\end{table}

%

\end{document}